\documentclass[sigconf]{acmart}

\AtBeginDocument{%
  \providecommand\BibTeX{{%
    \normalfont B\kern-0.5em{\scshape i\kern-0.25em b}\kern-0.8em\TeX}}}

\usepackage{listings}
\usepackage{array}
\usepackage{xspace} 
\usepackage[pdf]{graphviz}
\usepackage[normalem]{ulem} 
\usepackage{comment}
\usepackage{graphicx}
\graphicspath{ {./figures/} }
\usepackage{varwidth}
\usepackage[ruled, linesnumbered]{algorithm2e}
\usepackage{enumitem}
\usepackage{footnote}
\usepackage{multirow}
\usepackage{tablefootnote}
\usepackage{xpatch}
\usepackage{balance}
\usepackage{subcaption}
\usepackage{tabularx}

\copyrightyear{2022}
\acmYear{2022}
\setcopyright{acmcopyright}\acmConference[HILDA '22 ]{Workshop on
Human-In-the-Loop Data Analytics}{June 12, 2022}{Philadelphia, PA, USA}
\acmBooktitle{Workshop on Human-In-the-Loop Data Analytics (HILDA '22 ), June 12,
2022, Philadelphia, PA, USA}
\acmPrice{15.00}
\acmDOI{10.1145/3546930.3547495}
\acmISBN{978-1-4503-9442-0/22/06}

\begin{document}

\title{Towards Causal Physical Error Discovery in Video Analytics Systems}

\author{Ted Shaowang}

\authornote{Both authors contributed equally to this research.}
\email{swjz@uchicago.edu}
\orcid{0000-0002-7353-3786}
\author{Jinjin Zhao}
\authornotemark[1]
\email{j2zhao@uchicago.edu}
\affiliation{%
  \institution{University of Chicago}
  \country{USA}
  \postcode{60637}
}

\author{Stavros Sintos}
\email{sintos@uchicago.edu}
\affiliation{%
  \institution{University of Chicago}
  \country{USA}
  \postcode{60637}
}

\author{Sanjay Krishnan}
\email{skr@uchicago.edu}
\affiliation{%
  \institution{University of Chicago}
  \country{USA}
  \postcode{60637}
}

\renewcommand{\shortauthors}{Shaowang* and Zhao*, et al.}

\begin{abstract}
Video analytics systems based on deep learning models are often opaque and brittle and require explanation systems to help users debug. Current model explanation system are very good at giving literal explanations of behavior in terms of pixel contributions but cannot integrate information about the physical or systems processes that might influence a prediction. This paper introduces the idea that a simple form of causal reasoning, called a regression discontinuity design, can be used to associate changes in multiple key performance indicators to physical real world phenomena to give users a more actionable set of video analytics explanations. We overview the system architecture and describe a vision of the impact that such a system might have. 
\end{abstract}



\maketitle

\newcommand{\jj}[1]{\textcolor{red}{\bf [JJ: #1]}}

\newcommand{\sys}{VizEx\xspace}

\section{Introduction}
Advances in computer vision have led to a proliferation of \emph{video analytics} applications - from traffic pattern analysis to warehouse asset management.  
However, it is now an accepted reality that the deep learning models typically used can be brittle on real-world data~\cite{hendrycks2021natural}. This brittleness ranges from the obvious, such as camera occlusions and obstructions~\cite{chandel2015occlusion}, to the subtle, like natural adversarial examples where models are inexplicably fooled~\cite{hendrycks2021natural}. Every video analytics deployment will have inaccuracies that are attributable to one or more of such factors, and developers of today's video analytics systems have little recourse to determine why an error has occurred~\cite{ananthanarayanan2017real, viva, vocal, apostolo2022live, balayn2022can}. 

Video analytics inherently deals with our physical world and all of the complexity herein. 
In a sense, debugging video analytics is debugging the physical world.  
One needs to understand how real-world actions in space and time translate into the pixels captured by the camera.
Let’s consider an example where an object detection system is processing video from a fixed camera in a room to track the movement of furniture. 
Suppose that an occupant of the room turns off a light causing the scene to be obscured.
The consequence of such an action would be an abrupt drop in accuracy; however, many video analytics systems today would struggle to detect and attribute such a failure mode.
This is because debugging systems today rely on \emph{intrinsic explanations} where the failure is expected to be detectable from a model's direct inputs (i.e., the camera pixels) or outputs (i.e., detections from an object detection model).

\begin{figure}
    \centering
    \includegraphics[width=0.5\columnwidth]{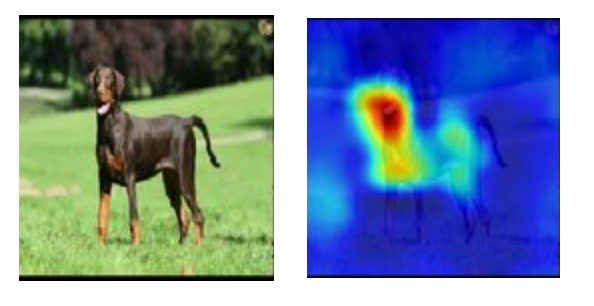}
    \caption{Saliency maps are a common explanation tool for computer vision models, which provide pixel-level contributions to a final output.}
    \label{fig:sal}
\end{figure}

Intrinsic explanations can sometimes fail to capture even simple failure modes like an abrupt lighting change.
For example, a standard algorithm in explainable video analytics (e.g., LIME~\cite{lime}) can tell an engineer which pixels influence a final prediction in a single frame (Figure \ref{fig:sal}). 
Since this explanation is itself high-dimensional, integrating these influence patterns over time can be noisy and unreliably. Correlating abrupt changes in the influence pattern to lighting (e.g., as opposed to object occlusion) can be very challenging if not impossible.
There are also scenarios where an engineer might only have access to prediction/explanation logs but not the actual source video, e.g., for privacy reasons~\cite{xu2019vstore}. In these cases, an engineer might understand which pixels contribute to a prediction but not what those pixels might represent. It might also be the case that the source video is available, but it is so diverse that semantically understanding what pixel positions correspond to in each video would be infeasible. 
In the above example, a much more straightforward approach could determine that lighting might have something to do with a reduction in accuracy. Let’s suppose that after every object detection prediction, we tracked the average luminosity~\cite{weeks1995edge} of each captured frame and the number of objects detected. The result of this tracking is a low-dimensional, bi-variate time series. At the time point where the light is turned off, we will observe a temporal discontinuity in the luminosity metric. Simultaneously, if we were to observe a drop in the number of objects detected at the point of discontinuity, there would be evidence that luminosity correlates with a drop in object detection accuracy. After discovering this correlation, human intervention can pinpoint that average luminosity is a proxy to lighting in the scenario by observing the video and applying knowledge about the physical world (i.e., an \emph{extrinsic explanation}). 

This example is a simple motivating scenario for this vision paper. We argue that purely using intrinsic explanations to debug large-scale video analytics deployments is not sustainable. 
A missing tool in video analytics is a framework that allows engineers to augment the video with low-dimensional key performance indicators (KPIs) derived from the video itself (e.g., luminosity), metadata (e.g., time of data), and external sensors (e.g., audio).
These low-dimensional signals that capture more domain knowledge of the physical world are used to explain changes in model behavior (e.g., a drop in the number of detected objects).

Since we are modeling multiple co-evolving and correlated time series, a key component of this framework is observational causal inference. 
One form of causal inference that can be employed is a regression discontinuity design (RDD)~\cite{hahn2001identification}. An RDD makes an underlying assumption that without an external impetus, an observed data relationship would be smooth. Likewise, the presence of a discontinuity would mean that some type of a causal intervention happened. By comparing observations lying on either side of the discontinuity, it is possible to estimate the effect of that intervention in environments where randomized experimentation is unfeasible. Video data naturally forms an RDD with a strong amount of smoothness in both time and pixel-space. Discontinuities in either axis can be used to identify potential causal relationships between KPIs about the predictions (e.g., the relationship between luminosity and number of detections).

We present a vision for a framework, \sys, that provides an engineer with a causal exploration of joint KPIs in video analytics systems. In our experiments, we use a set of KPIs that are proxies for common occurrences that happen in fixed-camera video analytics. Our framework tracks these KPIs over the course of a video and identifies discontinuities over frames in the KPI measurements. There is vast literature on observational causal inference in AI and Statistics~\cite{nichols2007causal}, and we use standard techniques from this literature to do our analysis. In fact, there is a growing body of work that studies causal inference in the context of databases~\cite{meliou2010causality, meliou2014causality}. To the best of our knowledge, this work has not been applied to debugging sensing problems or has leveraged RDDs. This paper presents a vision and initial results toward this goal. We describe the framework, the basic algorithms, and the  initial experiments that motivate our results.

\section{Motivation and Project Vision}
The confluence of large-scale machine learning and ubiquitous video capture promises a plethora of new applications. To make this vision more clear, we expand on the lighting example in the introduction.

\begin{example}\label{example}
In-home cameras can be used to recognize and track the activities done by occupants. 
Home activity recognition has applications in elder care, safety, and security. 
For privacy and complexity reasons, extracting significant amounts of labeled video from a particular home environment is mostly infeasible.
Thus, most such applications will likely rely on pre-trained activity recognition models deployed within a home's network.
By nature, home environments are highly unstructured -- no two homes and camera placements will be exactly comparable -- leading to widely varying activity recognition performance.
\emph{How can an engineer debug such a deployment without direct access to the video?}
\end{example}

Debugging questions for such applications could include:
\begin{enumerate}
    \item What types of camera angles lead to better or worse activity recognition?
    \item How does lighting affect activity recognition performance?
    \item What types of foreground objects affect performance?
\end{enumerate}
To answer such debugging questions, engineers need to be able to identify common failure patterns.

\subsection{Baseline Approach. Spatial Explanations}
Interpretable/explainable machine learning takes many forms, and we encourage the reader to refer to a recent textbook for a comprehensive survey~\cite{molnar2020interpretable}.
For computer vision applications, today's work can best be summarized as ``spatial explanations'', where the algorithms describe which pixels in an image influence a final prediction. 
Two examples of these techniques are LIME~\cite{lime} and saliency maps~\cite{saliencymap}.
Local interpretable model-agnostic explanations (LIME)~\cite{lime} is an algorithm to explain individual predictions of black-box models by approximating them locally with an interpretable model. With such an algorithm, we know how much each input feature would positively or negatively change the final prediction.
While very general, spatial explanations are high-dimensional and noisy, and additional modeling is needed to extract a real-world failure model from an anomalous sequence of explanations.

Another approach is to track features of the predictions the models make rather than the input pixels~\cite{kang2018model}.
While it might be more feasible to detect anomalies from such a framework, attribution to a real-world cause still requires significant human intervention.
In part, we argue that the problem setting in these video analytics debugging works is too complex for the current scale and scope of analytics deployments.
In the same way that programmers write logging utilities to detect known and common failure models in software systems, video analytics systems should have the same degree of human-designed logging.
In fixed-camera video deployments, we believe that it is not only feasible to enumerate signals that directly track many common sources of video analytics failures, but the responsible engineering solutions.
It is important to note that some failures may not have a software or modeling solution. For example, a problem with lighting or camera angle might require repositioning the camera.

\subsection{Our Approach. \sys}
We can much more readily capture properties of the physical world through programmed metrics (KPIs) that explicitly measure potential issues.
There are two general categories of KPIs that we can make use of: visual features and external data streams. Visual features include proxies for the debugging questions above, such as determining lighting through luminosity metrics, camera angle through depth modeling, and object composition by aggregating the detections. Extrinsic streams are temporally aligned with the video data, but come from other devices. For example, Wi-Fi network traffic data of smart home devices can help us determine which devices are currently interacting.
This stream gives us a coarse set of labels that can be easily leveraged without having to label a lot of video data manually.

Our goal is to elevate the level of abstraction in the logged explanations.
Our system, \sys, is motivated by the following insights.
\begin{enumerate}
    \item In video analytics deployments, there are simple-to-compute metrics that correlate with changes in prediction accuracy. These metrics can be proxies for lighting changes, camera angle movements, and disorder/occlusion in the scene.
    \item External sensors might be available and generate time series data temporally aligned with videos. They can be insightful when video analytics methods give uncertain or erroneous results.
    \item These metrics can be logged over frames of a video, creating a large, multivariate time series that is much smaller in dimensionality than a full saliency map.
    \item To account for spurious correlations, a regression discontinuity analysis can be performed locally to overlay a causal evidence structure over the collected metrics.
\end{enumerate}

\subsubsection{User-Defined Key Performance Indicators}
How does this work in practice? First, we ask users of \sys to define an evaluation metric on the performance of a model. This metric is task-specific and evaluates the general behavior of a model. For example, in an object detection task, they might count the number of objects detected in a frame. Or, in activity recognition tasks, we might track the average confidence value of activities detected over a time period.
Additionally, we have a set of KPIs that may have causal relationships with errors on the evaluation metric. KPIs can be simple, or they can be more complicated functions of a set of frames. In general, they will require little additional effort to compute.
The user defines this set of such KPIs based on their application and how much they would like \sys to centrally log the predictions.
A KPI $K_i(\lambda, w)$ applies the function $\lambda$ to windows of frames sized $w$, creating a time series that is streamed to a central analysis server.

\subsubsection{Query Processing}
Over such data, engineers can ask quasi-SQL queries of the form:
\begin{lstlisting}
SELECT *
FROM Video
WHERE metrics = 0
BECAUSE kpi_1 OR kpi_2

\end{lstlisting}

In the \texttt{WHERE} clause, engineers specify \texttt{metrics}, the desired evaluation metric over frames, and the value of \texttt{metrics} that correspond to the type of error they are interested in. In the motivating example of IoT devices, the engineers would be interested in activity recognition mispredictions. The \texttt{BECAUSE} clause specifies the KPIs. Examples of KPIs in the IoT example would include luminosity or camera angle. The query would return a sample of frames that have the desired evidence structure. 
In cases where the source video is not accessible, these SQL query results can be turned into aggregate statistics that illustrate relationships between the explanation metrics and the user-defined KPIs in real-world data.
In cases where the source video is accessible, these SQL query results can be used to retrieve frames for human inspection.

\subsection{Related Work}
Recent work explores the area of video analytics from different perspectives.
Some systems~\cite{blazeit, miris, lu2016optasia} specify query languages for fixed schemas, and others~\cite{rekall, scanner} makes it possible for users to specify ad-hoc queries over video. With \sys, we extend these types of query systems to include causality queries on the model.
Execution engines like Scanner~\cite{scanner} and VideoStorm~\cite{videostorm} manage hardware resources efficiently and run complex DNNs at scale.
To reduce the cost of running heavy neural networks on videos, model-level optimizations are developed to make predictions faster while preserving accuracy~\cite{kang2017noscope,focus}.
Other work~\cite{tasm, vstore} focuses on the storage and decoding of video data, which can be a bottleneck for video analytics as well.
~\cite{KazhamiakaZB21} envisions a new query system to address new challenges posed by autonomous vehicles (AV) data.

There have been traditional data explanation systems for structured data that perform feature selection and try to find the cause of error for users~\cite{bailis2016macrobase,diff,scorpion,dataxray,DBLP:conf/sigmod/RoyS14}.
More recent vision papers, such as VIVA~\cite{viva} and VOCAL~\cite{vocal} are interested in interactive video analysis. VIVA takes domain knowledge in structured format as user input and optimizes queries across unstructured and structured data. VOCAL tries to develop automatic ways of extracting and learning features. In the future, we aim to use these types of model-agnostic discovery systems to select complex input features.
\section{Case Study: VIRAT Dataset}\label{sec:analysis}
To illustrate the feasibility of this problem, a preliminary analysis was performed on the VIRAT video dataset~\cite{virat} to understand possible causal relationships in videos. This is a surveillance dataset that captures realistic human and vehicle behavior in an uncontrolled environment. We focused on frame-level person detection for three people-dominated scenes in the dataset (``0000'', ``0001'', ``0102''), and we debug the output of YOLOv3~\cite{yolov3}. 
One advantage of this dataset is the ground truth, so we can determine objectively how well our approach is doing.
For each frame, two categorical counting errors of the YOLO network were considered - whether the neural network undercounted or overcounted the number of people in the frame with bounding boxes.  Then, we show an example sequence that demonstrates the advantages of RDD to previous techniques. Finally, we illustrate the weaknesses of using other forms of explainable AI, such as fitting a decision tree. We frame this case study as answering some basic hypotheses to illustrate the value of a system like \sys.

\begin{figure*}[t]
\begin{subfigure}[h]{0.3\linewidth}
\includegraphics[width=\linewidth]{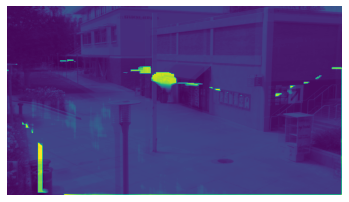}
\caption{Overcounted}
\end{subfigure}
\hfill
\begin{subfigure}[h]{0.3\linewidth}
\includegraphics[width=\linewidth]{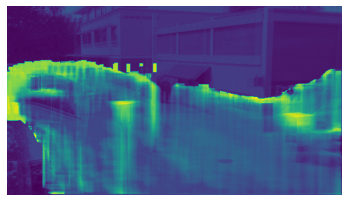}
\caption{Undercounted}
\end{subfigure}
\hfill
\begin{subfigure}[h]{0.3\linewidth}
\includegraphics[width=\linewidth]{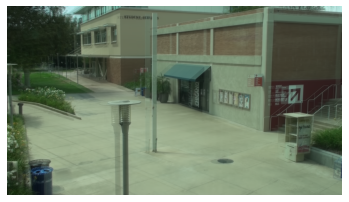}
\caption{Original}
\end{subfigure}
\caption{YOLOv3 undercounts and overcounts ``people'' in very specific areas of a scene corresponding to occlusion, disorder, or lighting.}
\label{fig:0102}
\end{figure*}

\subsection{Motivating Results}
\emph{Hypothesis 0. There are consistent spatial and temporal patterns in prediction errors in video analytics tasks. } The errors that deep learning models make are not as unpredictable as one would think. In our first hypothesis, we illustrate how over the course of a long video, there are clear patterns in where and how errors occur. Figure ~\ref{fig:0102} illustrates the accuracy of YOLOv3 in different parts of a fixed-camera scene. We use the ground truth annotations to determine in which areas the count of the number ``people'' is most accurate. 
Figure ~\ref{fig:0102}(a) shows a heatmap of areas where YOLOv3 tends to over-count people, Figure ~\ref{fig:0102}(b) shows the same for under-counting. Not surprisingly, over-counting correlates with parts of the scene with a high number of objects and sharp edges. Under-counting correlates with occlusion and lighting. For example, we see a street light in green color, which obstructed people walking behind it and causes false negatives.

A significant takeaway of these figures is that, while we currently do not understand the inner mechanics of deep learning models, there do seem to be patterns in errors that are linked to real-world visual concepts in the video. Humans can have intuitive estimations of causality from those patterns, given sufficient evidence. However, only relying on human intuition can lead to bias and a lack of quantitative evaluation. It may also be difficult for humans to parse the video for patterns (for example, non-static videos would not have such clean spatial heatmaps). Hence, in \sys, we would like to integrate both automatic detection methods and human intuition verification to discover causality from error patterns in the video.

\subsection{Discontinuity Analysis}
\emph{Hypothesis 1. It is possible to use KPIs to generate causal explanations for spatially and temporally correlated errors. }
Discontinuity analysis isolates properties that change when errors occur. In this section, we give an example sequence of frames where low-level discontinuity corresponds to causal errors in VIRAT. We demonstrate how an RDD might be able to generate such explanations. For this experiment, we leverage a ground truth scene that shows a group of people walking. We track the luminosity around the detected objects as a function of time. We also track the number of objects detected. 
Figure~\ref{fig:luminosity} shows the average luminosity drops (left y-axis, colored in blue) and the percentage of frames with correct counts also drops (right y-axis, colored in green). This corresponds to people walking toward a dark area in the frame. Two vertical lines correspond to the midpoints B and C in figure~\ref{fig:screenshots}.  As people walk into the dark area (figure~\ref{fig:screenshots}a-d), the luminosity drops and there is a significant increase in undercounting errors (decrease in correct counts). Discontinuity analysis allows us to collect evidence that luminosity may be related to the increase in under-counting errors because there is a simultaneously spatial and temporal change in these metrics.

\begin{samepage}
\begin{figure}[t]
\includegraphics[width=0.8\linewidth]{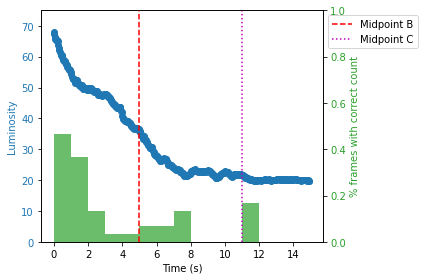}
\caption{Average luminosity value \& accuracy for each frame}
\label{fig:luminosity}

\vspace*{\floatsep}

\begin{subfigure}[h]{0.22\linewidth}
\includegraphics[width=\linewidth]{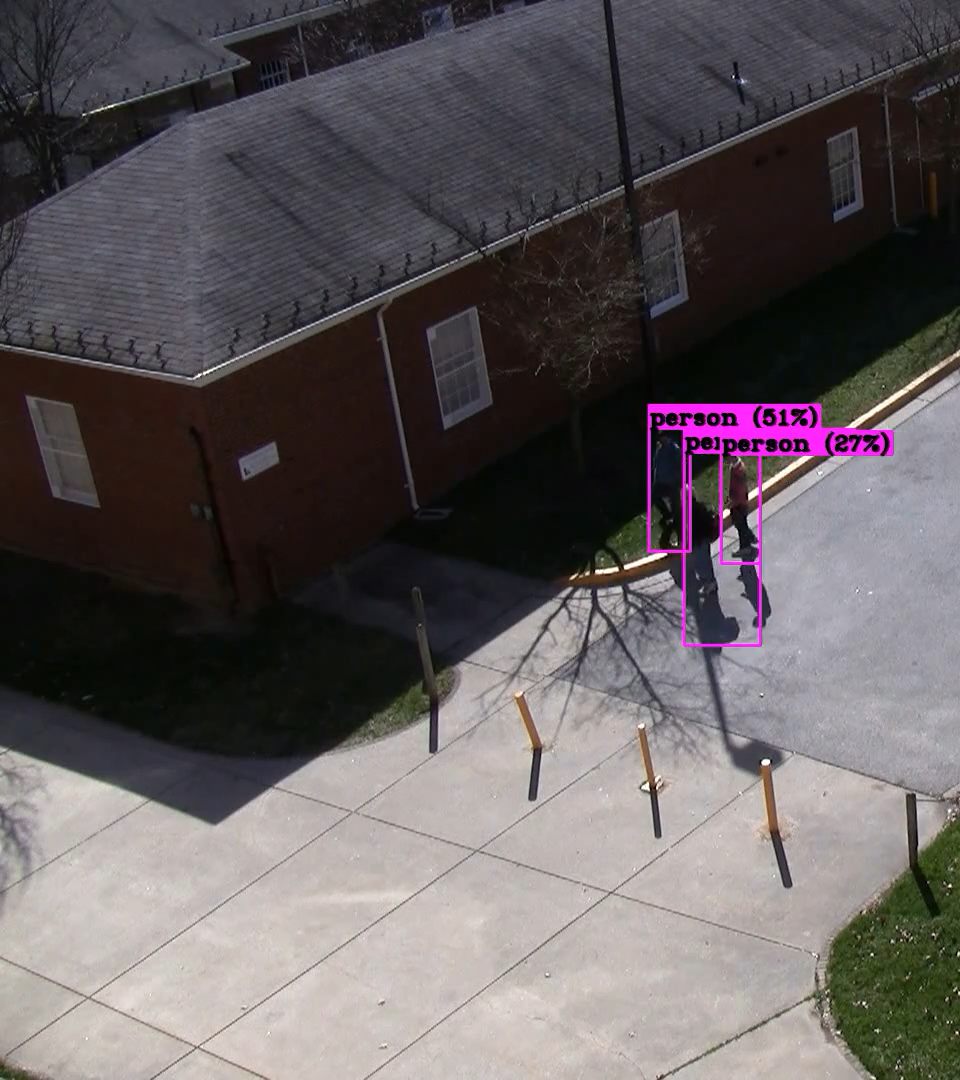}
\caption{Start}
\end{subfigure}
\hfill
\begin{subfigure}[h]{0.22\linewidth}
\includegraphics[width=\linewidth]{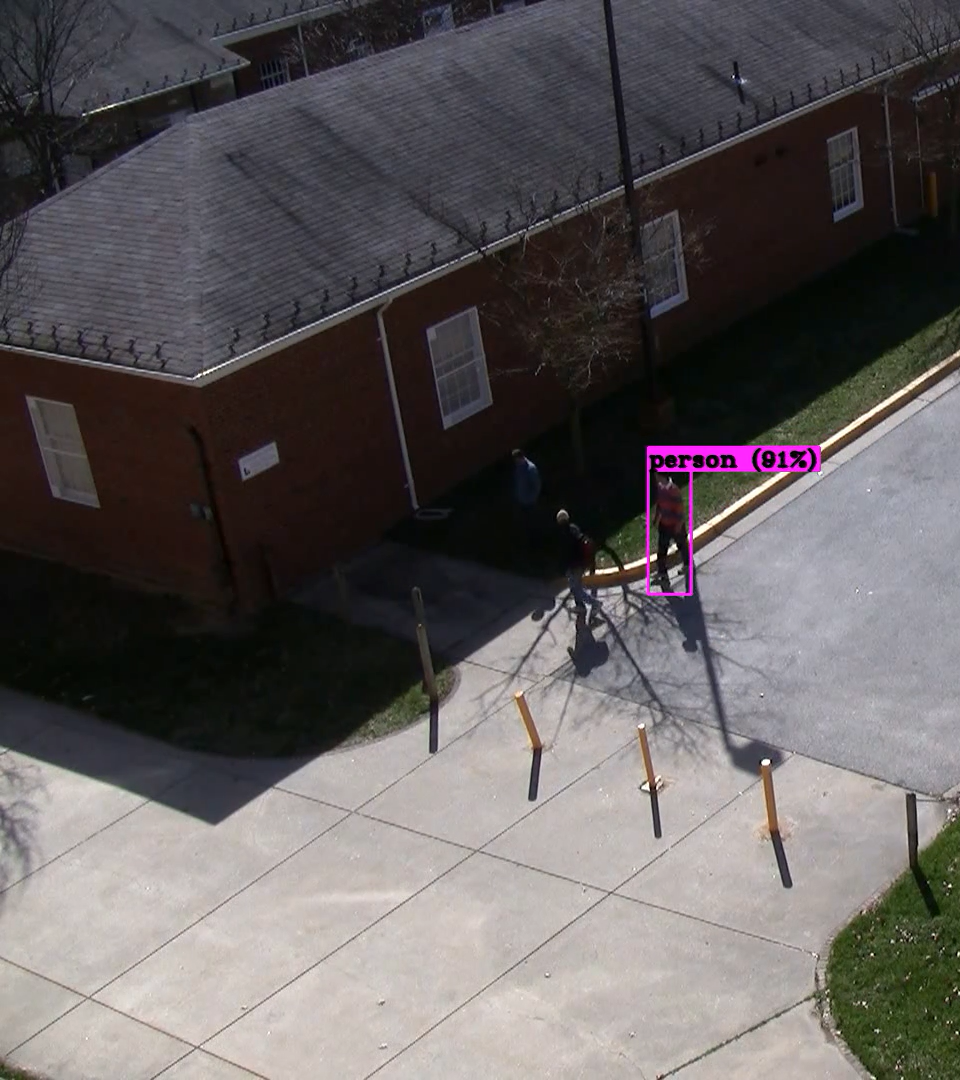}
\caption{Midpoint B}
\end{subfigure}
\hfill
\begin{subfigure}[h]{0.22\linewidth}
\includegraphics[width=\linewidth]{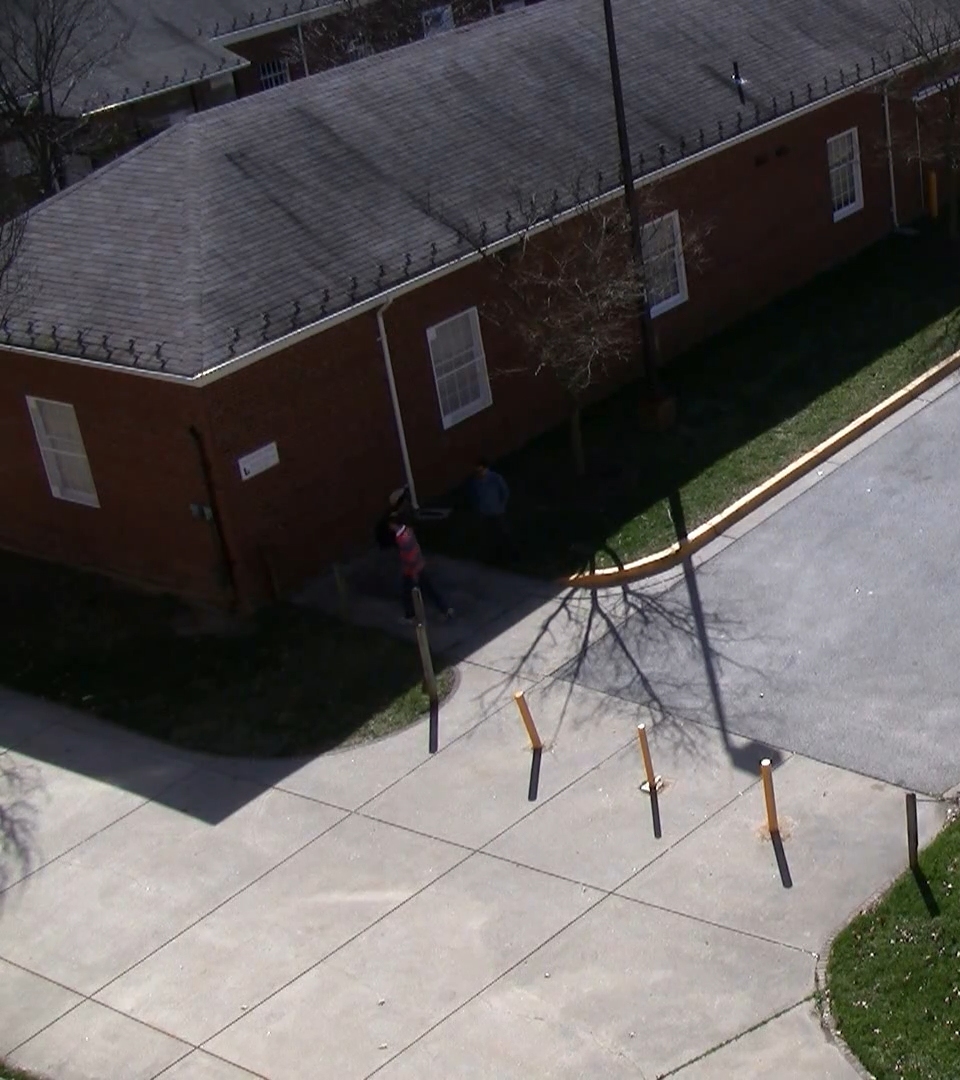}
\caption{Midpoint C}
\end{subfigure}
\hfill
\begin{subfigure}[h]{0.22\linewidth}
\includegraphics[width=\linewidth]{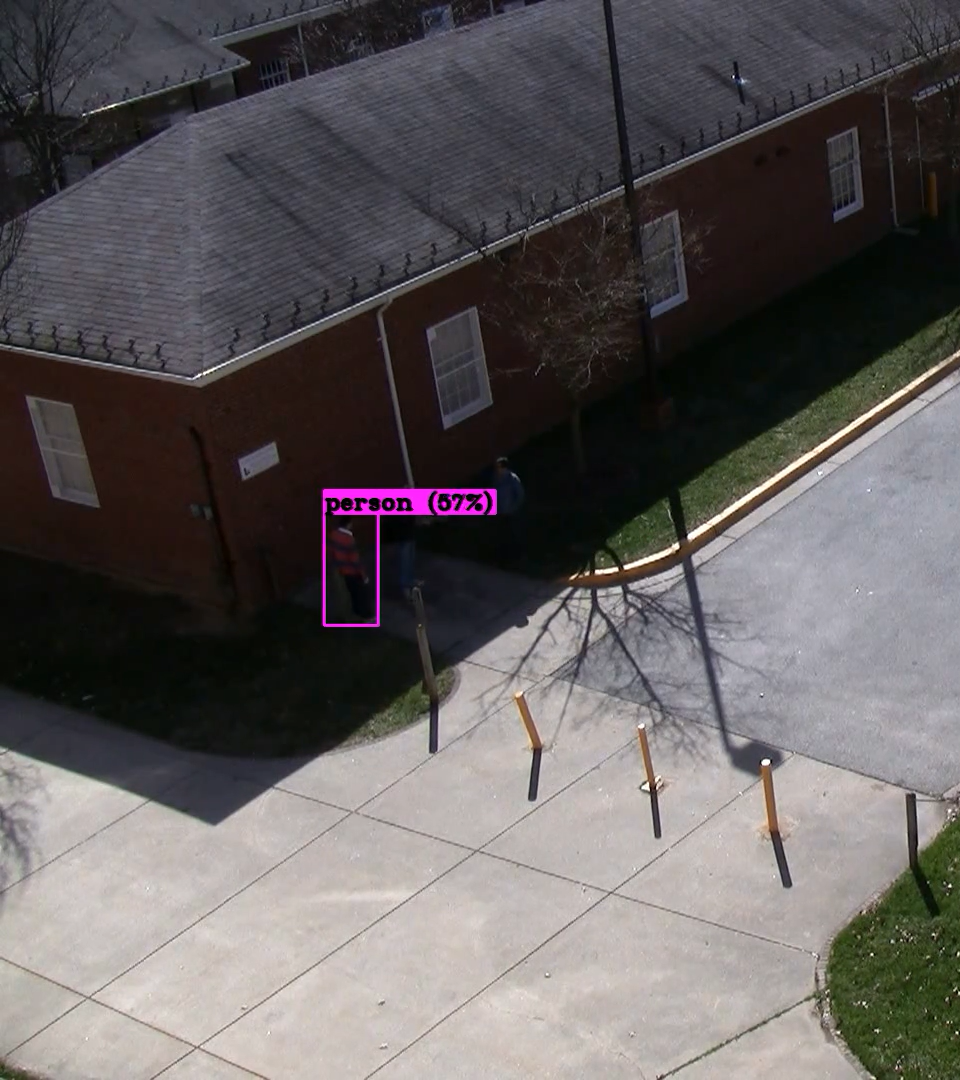}
\caption{End}
\end{subfigure}
\caption{Still images from the scene corresponding to figure~\ref{fig:luminosity}}
\label{fig:screenshots}

\end{figure}
\end{samepage}

\subsection{Decision Trees As An Alternative}
\emph{Hypothesis 2. Surrogate Model Explanations Are Insufficient For Video Analytics Debugging.} A popular explainable AI technique is to use a surrogate model, which is a simplified model that is human-interpretable that mimics the behavior of a more complex model (at least locally).  We find that such an approach is difficult to operationalize in this setting. There are simply too many spurious correlations to train an accurate surrogate. The RDD approach is a principled way of rejecting/controlling for spurious correlations by only looking for simultaneous changes. To illustrate this, we run some initial experiments using a decision tree to explain the relationship between the KPIs.

\begin{table}[h]
    \centering
    \begin{tabularx}{.9\columnwidth}{lll}
    \hline
         & Balanced & Balanced \\
        Dataset & Training Accuracy & Testing Accuracy \\
         & (\%) & (\%) \\
    \hline
        ``0000" & 62.24 & 32.90 \\
        ``0001" & 64.20 & 31.97 \\
        ``0102" & 48.66 & 33.72 \\
        ALL & 80.74 & 34.05 \\
    \end{tabularx}
    \caption{A table showing balanced accuracy metrics of decision trees on error detection}
    \label{table:decision-tree}
\end{table}



We modeled a decision tree with a depth of 10 on each individual scene and across all three scenes, and evaluated the results. The evaluation metric was the accuracy in predicting three classes: correct counts, under-counting, or over-counting. The training and testing datasets for individual scenes were split evenly, and we used the first two scenes as the training dataset and the last scene as the testing dataset when all scenes were evaluated together. For all datasets, we sampled one frame per second for each video. We chose to evaluate 5 different features in this analysis - average color (3 RGB channels), luminosity, and percentage of edges (with Canny edge detector). These features were calculated for the whole image, for a 4x4 grid for each frame. The average of these features for all true-labeled and YOLO-detected bounding boxes was also calculated. These values were used as the input features of the decision trees.

Table \ref{table:decision-tree} shows the balanced accuracy of each decision over the training and testing dataset. As we see in the table, trends are consistent across multiple datasets. The decision tree provides reasonable accuracy on the training dataset, but doesn't predict better than random on the testing dataset, even within the same scene. This suggests that the decision tree is learning local correlations but really cannot generalize to unseen examples. We strongly hypothesize that these local correlations are highly dependent on spatial and temporal closeness - i.e., our model predicts bounding boxes that are close spatially and temporally as being similar.

\section{Discussion}

We introduced a novel issue of causality reasoning over video analytic models. Preliminary analysis suggests that potential causal relationships could exist between low-level KPIs and model errors, and decision tree models are not sufficient for capturing those relationships. We presented the idea of RDD, as an alternative to those relationships, and outlined the high-level design of \sys, a causal query system. In the future, we aim to build \sys, and fully evaluate the concepts envisioned in this paper.

Given a reasonable definition of KPIs and desired metrics, our system is able to attribute observed errors to common patterns without access to original videos. Back to Example~\ref{example}, we would be able to debug an activity recognition pipeline by logging explanations at the same time each prediction is made. Those explanations would be human-readable and engineers no longer have to watch hours of videos in order to find the cause of such errors, let alone privacy concerns. They can adjust the full pipeline to deal with the errors; e.g. if they found camera angle causality, they can have a pre-processing step that digitally adjusts the image, or if they found obstacle causality, they can specify in the camera setup to make a clear path for the camera.

\subsection{Integration of RDD}
There will need to be multiple adjustments to the base RDD concept in order to fit the video use case. Firstly, as seen in Figure~\ref{fig:screenshots}, error labels on the data are not clean, and we will need a careful definition of boundaries for when error sequences occur. In the video, the event of the user entering the shadow area is not instantaneous. Secondly, we want to compare KPIs across multiple error sequences and RDD results in order to validate causation over multiple scenes and cluster sequences with similar causes. We aim to study and experiment with the optimal method for this in the future.

\subsection{User Interactivity}
We believe that RDD is useful for the discovery of KPIs that are strongly correlated with true physical causal indicators. However, our algorithm would still ultimately rely on the user to link KPIs to those physical indicators. This user interaction occurs at two stages - when the user selects the task-appropriate KPIs for querying, and when the user evaluates the final results of the queries. In order to facilitate the second state, we expect \sys to present useful information in the query results. We envision a combination of example sequences, RDD plots and summary statistics. Presenting this information isn't trivial since video datasets can be very large, and can contain many edge cases \cite{vocal}. \sys cannot visualize every relevant error sequence, but the user may make an incorrect decision if the wrong ones are chosen. 

\subsection{The Presence of Ground Truths}
Most of our preliminary analysis in Section~\ref{sec:analysis} assumes the presence of ground truths, but that is not always the case in real-world scenarios. While true labels make such analysis easier, having them is not a requirement of our system. Our definition of the evaluation metric does not require ground truths as long as the user defines an appropriate \texttt{labels} metric that applies to windows of frames. Understanding how types of metrics differ may allow us to use different techniques to assess correlation.

\balance

\bibliographystyle{ACM-Reference-Format}
\bibliography{reference}
\end{document}